\documentclass[11pt]{article}

\usepackage[final]{acl}

\usepackage{times}
\usepackage{latexsym}

\usepackage[T1]{fontenc}

\usepackage[utf8]{inputenc}

\usepackage{microtype}

\usepackage{inconsolata}

\usepackage{graphicx}

\usepackage{algorithm}
\usepackage{algpseudocode}
\usepackage{amsmath,amssymb}

\usepackage{tcolorbox}
\usepackage{booktabs}
\usepackage{multirow}
\usepackage{multicol}
\usepackage{pifont}
\usepackage{bytefield}
\usepackage{tikz}
\usepackage[table]{xcolor}

\usepackage[shortlabels,inline]{enumitem}
\usepackage{tikz,lipsum}
\usepackage{ragged2e}
\usepackage[dvipsnames]{xcolor}
\usepackage{amsmath}
\usepackage{booktabs}
\usepackage{tabularray}
\usepackage{arydshln}
\usepackage{stmaryrd}
\usepackage{marvosym}
\usepackage{colortbl}
\usepackage{multicol}
\usepackage{multirow}
\usepackage{float}
\usepackage{amsfonts}
\usepackage{amssymb}

\usepackage{cleveref}
\crefname{section}{\S}{\S}
\crefname{table}{Table}{Tables}
\crefname{figure}{Fig.}{Figs.}
\crefname{algorithm}{Alg.}{}
\crefname{ALC@unique}{Line}{Lines}
\crefname{equation}{Eq.}{Eqs.}
\crefname{appendix}{App.}{Apps.}
\crefformat{section}{\S#2#1#3}

\usepackage{soul}
\definecolor{tablegray}{RGB}{223, 242, 252}

\usepackage{todonotes}

\NewDocumentCommand{\prompt}{O{} +m}{%
\begin{tcolorbox}[
    coltitle=white,
    colframe=black,
    colback=black!5!white,
    boxrule=1pt,
    enhanced jigsaw,
    breakable,
    pad at break*=2mm,
    left=2pt,
    right=2pt,
    top=2pt,
    bottom=2pt,
    fontupper=\small,
    fontlower=\small,
    title={#1}, 
]
#2 
\end{tcolorbox}
}

\title{BanglaTalk: Towards Real-Time Speech Assistance for Bengali Regional Dialects}

\author{Jakir Hasan \\
  Shahjalal University of Science\\
  and Technology, BD \\
  \texttt{jakirhasan718@gmail.com} \\\And
  Shubhashis Roy Dipta \\
  University of Maryland, Baltimore\\ 
  County, USA \\
  \texttt{sroydip1@umbc.edu} \\}

\begin{document}
\maketitle

\begin{abstract}


Real-time speech assistants are becoming increasingly popular for ensuring improved accessibility to information. Bengali, being a low-resource language with a high regional dialectal diversity, has seen limited progress in developing such systems. Existing systems are not optimized for real-time use and focus only on standard Bengali. In this work, we present \textbf{BanglaTalk}, the first real-time speech assistance system for Bengali regional dialects. BanglaTalk follows the client-server architecture and uses the Real-time Transport Protocol (RTP) to ensure low-latency communication. To address dialectal variation, we introduce a dialect-aware ASR system, \textbf{BRDialect}, developed by fine-tuning the IndicWav2Vec model in ten Bengali regional dialects. It outperforms the baseline ASR models by 12.41-33.98\%
on the RegSpeech12 dataset. Furthermore, BanglaTalk can operate at a low bandwidth of 24 kbps while maintaining an average end-to-end delay of 4.9 seconds. Low bandwidth usage and minimal end-to-end delay make the system both cost-effective and interactive for real-time use cases, enabling inclusive and accessible speech technology for the diverse community of Bengali speakers.\footnote{\url{https://github.com/Jak57/BanglaTalk}}




\end{abstract}

\section{Introduction}


\begin{figure}[!t]
    \centering
    \begin{tikzpicture}
        \node[rounded corners=20pt, draw=none, inner sep=0pt, outer sep=0pt, clip] 
            {\includegraphics[width=\linewidth]{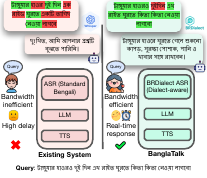}};
    \end{tikzpicture}
    \caption{Existing Bengali speech assistants (left) fail to understand queries in regional dialects due to reliance on standard Bengali ASR (incorrect transcriptions are shown in red). \textbf{BanglaTalk (right) successfully handles regional dialect queries through its dialect-aware ASR (BRDialect).} It is bandwidth efficient and operates in real-time due to the incorporation of the Real-Time Transport Protocol.}
    \label{fig:intro}
\end{figure}

Conversational speech assistants \citep{dutsinma2022systematic} have transformed human-computer interaction, making information more accessible. Widely adopted tools such as Alexa, Siri, and Cortana demonstrate the profound impact of real-time speech assistants on human lives \citep{hoy2018alexa}. However, while significant progress has been made for high-resource languages such as English, Mandarin, and French, such tools are still underdeveloped for the low-resource Bengali language. Bengali is a morphologically rich Indo-Aryan language \citep{islam2025banglalem}, spoken by approximately 260 million people worldwide.  It exhibits significant regional dialectal diversity, with variations in phonology, vocabulary, and syntax \citep{hasan2024credibility}. This linguistic diversity poses a major challenge in building robust speech assistants. 

Automatic Speech Recognition (ASR) is a key component of speech assistant systems. Existing ASR systems are developed primarily for standard Bengali \citep{saha2021development, rakib2023bangla}, and their performance is significantly degraded in regional dialects. As a result, existing speech assistant systems that integrate such ASR cannot support regional dialectal communication \citep{hasan2021alapi, arnab2023shohojogi}. Moreover, real-time deployment requires not only dialectal robustness, but also minimal end-to-end delay and efficient bandwidth usage. Previous works lack dialect-aware ASR, systematic analysis of delay minimization techniques, bandwidth efficiency, and real-time communication.

In this work, we introduce BanglaTalk, the first real-time conversational speech assistant for Bengali regional dialects. BanglaTalk adopts a client-server architecture and incorporates the Real-time Transport Protocol (RTP) \citep{schulzrinne2003rtp} to achieve low-latency communication. Robust audio encoding enables operation at 24 kbps (kilobits per second). As illustrated in \cref{fig:intro}, while existing speech assistants fail in interpreting regional dialectal queries, BanglaTalk transcribes them accurately through the dialect-aware ASR system. It responds to queries effectively and interactively in real-time using low bandwidth.

The BanglaTalk client integrates lightweight audio processing modules, including noise cancellation, dynamic range compression, and audio encoding. On the server side, a dialect-aware ASR system, a voice activity detector (VAD), a natural-sounding Text-to-Speech (TTS) system, and audio encoding modules form a complete pipeline for real-time speech assistance. Central to this system is BRDialect, a dialect-aware ASR model fine-tuned on ten Bengali regional dialects. BRDialect outperforms the baseline Whisper \citep{BengaliAIASRSubmission2023} and IndicWav2Vec \citep{javed2022towards} models, achieving a word error rate of 74.1\% and character error rate of 40.6\% on the RegSpeech12 \citep{regspeech12} dataset. 

Additionally, the integrated VITS \citep{kim2021conditional} TTS model produces natural-sounding speech with a high mean opinion score (MOS) of 4.49, enhancing the user experience.
With an average end-to-end delay of 4.9 seconds, BanglaTalk enables interactive real-time communication between the user and the speech assistant. This system will significantly impact the lives of Bengali speakers due to its dialect-aware ASR, low bandwidth usage, and real-time performance.

In summary, our main contributions are:
\begin{itemize}
    \item We introduce BanglaTalk, the first real-time, bandwidth-efficient Bengali speech assistant designed to support regional dialects through a client-server architecture.
    
    \item We develop BRDialect, a dialect-aware ASR system that substantially outperforms existing ASR models on the RegSpeech12 dataset spanning twelve regions of Bangladesh. 

    \item We provide a comprehensive analysis of audio processing latency, bandwidth usage, end-to-end delay, and generated speech quality, demonstrating the robustness of BanglaTalk for real-time, dialect-aware communication.
    
\end{itemize}

\section{Methodology}

\begin{figure*}[!t]
    \centering
    \begin{tikzpicture}
        \node[rounded corners=20pt, draw=none, inner sep=0pt, outer sep=0pt, clip] 
            {\includegraphics[width=\linewidth]{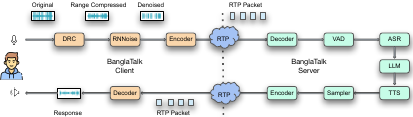}};
    \end{tikzpicture}
    \caption{Client (left) and server-side (right) processing pipelines of the BanglaTalk System.}
    \label{fig:client_server}
\end{figure*}

\textbf{BanglaTalk} follows a client-server architecture, with lightweight audio processing on the client and computationally intensive tasks on a centralized server. The overall pipeline is illustrated in \cref{fig:client_server}.

\subsection{BanglaTalk Client}
The client is responsible for capturing, processing, and transmitting audio to the server. As shown in \cref{fig:client_server} (left), its main modules include audio capture, dynamic range compression, noise suppression, encoding, and transmission.

\paragraph{Audio Capture} \label{sec:capture_audio}
The client captures audio in 20-ms (milliseconds) frames at a sample rate of 16 kHz (kilohertz). Each frame contains 320 samples in 16-bit PCM little-endian format \citep{dobson2000developments}. Although the Opus codec \citep{valin2016high} supports multiple sample rates (e.g., 8-48 kHz), we fix the sample rate at 16 kHz to align with the ASR and VAD modules. Opus allows frame durations of 2.5-100 ms. A frame duration of 20 ms (50 RTP packets per second) offers a balance between packet size and loss rate in real-time communication.

\paragraph{Dynamic Range Compression (DRC)} \label{sec:drc}
Speech captured from the microphone often includes soft and excessively loud signals. Compressing the dynamic range helps maintain a consistent audio level, enhancing performance \citep{giannoulis2012digital}. We develop the dynamic range compression algorithm described in \cref{alg:drc} and apply compression only to the loud audio segments. Whenever the decibel level of a normalized audio sample exceeds -10 dBFS (decibels relative to full scale), we apply a compression ratio of 2:1. Samples outside this threshold remain unchanged.

\paragraph{RNNoise Cancellation} \label{sec:noise_cancel}
Noise poses a major issue in audio communication. Background and foreground noise can be picked up by the microphone and transmitted to the network, degrading the overall performance of the system. To mitigate this, we perform noise suppression with RNNoise \citep{valin2018hybrid}. It is a lightweight neural network-based denoiser capable of real-time operation.

RNNoise is trained to remove noise from an audio frame of duration 10 ms at a 48 kHz sample rate. Since every captured audio frame duration is 20 ms at a 16 kHz sample rate, we apply audio segmentation, upsampling, and downsampling during noise suppression. \cref{alg:upsample} shows the pseudocode for upsampling. Linear interpolation \citep{xu2022research} is used in upsampling from 16 to 48 kHz due to minimal computational overhead. After denoising, the audio is downsampled to 16 kHz using \cref{alg:downsample}. Specifically, we skip intermediate sample values -- from every three consecutive audio samples, the first one is retained and the remaining two are discarded. This simple technique is computationally efficient for downsampling.

\paragraph{Encoding with Opus Codec} \label{sec:encoding_opus}
Opus is a high-quality audio codec for interactive speech and music transmission over the Internet \citep{valin2016high}. It is widely used in VoIP (Voice over IP) applications \citep{sundvall2014opus} due to the low latency processing and error concealment. Each audio frame contains a total of 320 samples (640 bytes). Without compression, audio frames are expensive to transmit over the Internet due to high bandwidth consumption. To mitigate this, each audio frame is encoded using the Opus codec with a low bitrate of 24 kbps. This ensures low bandwidth usage, which is a critical factor for greater accessibility. The incoming audio frames from the server are decoded with the same codec.

\paragraph{Packetization and Transmission} \label{sec:packet}
Each encoded audio frame is encapsulated in an RTP packet following the Real-time Transport Protocol \citep{audio1996rfc1889}. The first 12 bytes of each packet contain the header, and the remaining bytes contain the Opus-encoded payload. \cref{sec:app_rtp_structure} describes the structure of the RTP packet. These packets are sent at 20 ms intervals to the server's public IP and the RTP packet receiver port \citep{postel1980user}.

\subsection{BanglaTalk Server}
The server is responsible for receiving the audio stream from the client and generating an appropriate audio response. The overall server-side processing pipeline is illustrated in \cref{fig:client_server} (right). To function effectively, the server employs several interconnected modules. First, the incoming audio frames are received and decoded using the Opus decoder. Next, voice activity detection is performed to identify speech segments. When a complete user query is detected, the corresponding speech segment is transcribed into text. This text is then processed by a large language model (LLM), which generates a suitable response. The response is subsequently converted into speech using a TTS system. Finally, the speech is resampled and the resulting audio frames are encoded with the Opus codec. Following the RTP protocol, the created RTP packets are transmitted to the client. The detailed workflow of these modules is described below.

\paragraph{RTP Packet Parser}
The server receives audio data from the client in the form of RTP packets. Each packet is parsed to extract header information, encoded data length, and encoded audio data in byte format.

\paragraph{Decoding with Opus Codec}
The extracted encoded audio is decoded using the Opus codec. Based on the encoded data length and audio bytes, the decoder reconstructs audio frames of 20 ms duration, corresponding to 320 samples at a 16 kHz sample rate.

\begin{algorithm}[!t]
\caption{DRC compresses the amplitude of samples exceeding a $-10$ dBFS threshold by applying a 2:1 compression ratio. It leaves the quieter samples unchanged.}
\label{alg:drc}
\begin{algorithmic}[1]
\Require Frame $x \in \mathbb{Z}^{N}$, threshold $\tau=-10$ dBFS, ratio $r=2$
\Ensure  Compressed frame $y \in \mathbb{Z}^{N}$ 
\State $y \gets x$
\For{$i \gets 1$ to $N$}
  \State $s \gets y_i$
  \If{$s = 0$} \State \textbf{continue} \EndIf
  \State $d \gets 20 \cdot \log_{10}\!\big(\,|s|/32768\,\big)$ 
  \If{$d \le \tau$} \State \textbf{continue} \EndIf
  \State $d' \gets \tau + (d - \tau)/r$ 
  \State $\sigma \gets \mathrm{sign}(s)$ 
  \State $s' \gets \left\lfloor 10^{\,d'/20} \cdot \sigma \cdot 32767 \right\rfloor$
  \State $y_i \gets s'$
\EndFor
\State \Return $y$
\end{algorithmic}
\end{algorithm}

\paragraph{Voice Activity Detection (VAD)}
Voice activity detection is a crucial component of real-time communication, as it identifies speech segments while discarding non-speech portions. This prevents unnecessary downstream processing, thus reducing latency and improving overall efficiency. We use the Silero VAD \citep{SileroVAD} in the streaming mode, which is specifically designed for real-time applications. It works with a frame duration of 32 ms (512 samples) at a sample rate of 16 kHz and determines whether each audio frame marks the beginning, end of a speech segment, or none. Only the detected speech segment corresponding to the user query is forwarded to the ASR system.

\paragraph{Automatic Speech Recognition (ASR)}
We train a Wav2Vec2-based model \citep{baevski2020wav2vec} using speech data from ten regional dialects from the Ben10 dataset \citep{ben10}. For data processing, we follow \citet{hasan_saim_mostafa_2024_bengaliASR_indicwav2vec} and fine-tune the pre-trained IndicWav2Vec \citep{javed2022towards} for Bengali on that processed dataset. To evaluate the performance of our trained ASR model, we use the RegSpeech12 \citep{regspeech12} test set (Ben10 test set is not publicly available). A detailed description and analysis of these datasets are provided in \cref{sec:app_ben10} and \cref{sec:app_regspeech12}.

\paragraph{End of Query Detection}
Accurate detection of the end of a user query is essential for real-time speech assistance systems \citep{liang2023dynamic}. We have defined the end of query as a silence segment lasting at least 1.2 seconds. For silence detection, Silero VAD is utilized. Once an end-of-query is detected, the speech segment is passed to the ASR system to generate the transcription. The resulting query is then forwarded to the LLM, which produces the system's response.

\paragraph{Generating Response Using LLM}
Large Language Models (LLMs) are crucial for generating responses to user queries \citep{dam2024complete}. To maintain coherent communication, responses are generated for valid queries, while invalid queries are discarded. We employ \texttt{GPT-4.1-nano} as the chat model, with the prompt template presented in \cref{sec:llm_prompt}. To minimize latency, we use streaming mode, which delivers responses incrementally rather than waiting for a full response. The streamed text is segmented based on Bengali punctuation and forwarded to the TTS system.

\paragraph{Text-to-Speech (TTS)}
Several TTS models are available for Bengali \citep{raju2019bangla}. We experiment with MMS-TTS-Ben \citep{pratap2024scaling} and two variants of VITS-Bengali (male and female voices) \citep{ComprehensiveBanglaTTS2023}. These models are selected because of their minimal processing delay. MMS-TTS-Ben produces speech at a 16 kHz sample rate, while VITS-Bengali outputs at 22.05 kHz. For system compatibility, the VITS-generated speech is resampled to 16 kHz.

\paragraph{Network Transmission}
The processed audio is segmented into 20 ms frames and encoded with the Opus codec at a bitrate of 24 kbps. Each RTP packet is constructed with a 12-byte header, followed by encoded audio data. The RTP packets are transmitted over the Internet to the client's public IP and port at 20 ms intervals, ensuring synchronized real-time playback.

\section{Result \& Discussion}

\subsection{Implementation Details}
Since the system is intended for deployment across a diverse population in Bangladesh, we prioritize computational efficiency on the client side to ensure accessibility across devices with varying hardware capabilities. In contrast, the server must be sufficiently powerful to handle audio processing, response generation, and real-time communication with minimal latency. Accordingly, our experiments were conducted with an Intel Core i7 CPU (without GPU) as the client and a server equipped with an NVIDIA GeForce RTX 4090 GPU for efficient processing.

\begin{table}[!t]
\centering

\begin{tabular}{@{}lcc@{}}
\toprule
\textbf{Model} & \textbf{WER} $\downarrow$ & \textbf{CER} $\downarrow$\\
\midrule
Whisper-medium-Bengali & 0.846 & 0.562\\
IndicWav2Vec-Bengali & 0.897 & 0.615\\
BRDialect & \textbf{0.741} & \textbf{0.406}\\
\bottomrule

\end{tabular}
\caption{Performance of ASR systems on the test set of the RegSpeech12 dataset, covering twelve Bengali regional dialects.}
\label{tab:stt_performance}
\end{table}



\subsection{Evaluating BRDialect}
To evaluate the performance of our dialect-aware ASR system, \textbf{BRDialect}, we use the test set from the RegSpeech12 \citep{regspeech12} dataset. It includes dialects from twelve regions of Bangladesh -- Rangpur, Sylhet, Chittagong, Noakhali, Narail, Kishoreganj, Barishal, Habiganj, Comilla, Tangail, Sandwip, and Narsingdi, totaling 2132 audio files.

For evaluation, we report the Word Error Rate (WER) and Character Error Rate (CER), following the formulas described in \cref{sec:wer_cer_formula}. To refine ASR predictions, we apply beam search decoding with a 5-gram KenLM language model \citep{heafield2011kenlm}. We also investigated the impact of the preprocessing and postprocessing steps, including noise cancellation, normalization, and punctuation removal, as these factors significantly affect the overall performance of the ASR.

As shown in \cref{tab:stt_performance}, \textbf{BRDialect} outperforms baseline models in both WER and CER. Specifically, it achieves the lowest WER of 0.741 and CER of 0.406 when decoded with a 5-gram KenLM model, combined with Unicode normalization and punctuation removal, but without noise cancellation.

BRDialect consistently outperforms \texttt{Whisper-medium-Bengali} \citep{BengaliAIASRSubmission2023} and \texttt{IndicWav2Vec-Bengali} \citep{javed2022towards}, achieving a {12.41–17.39\%} relative improvement in WER and a {27.77–33.98\%} improvement in CER. The comparatively poor performance of the baseline models suggests that dialectal variation is not adequately addressed during their training, limiting their suitability for regional speech-to-text tasks. BRDialect highlights the importance of dialect-aware fine-tuning for building a robust ASR system.

\begin{figure}[!t]
    \centering
    \includegraphics[width=1\linewidth]{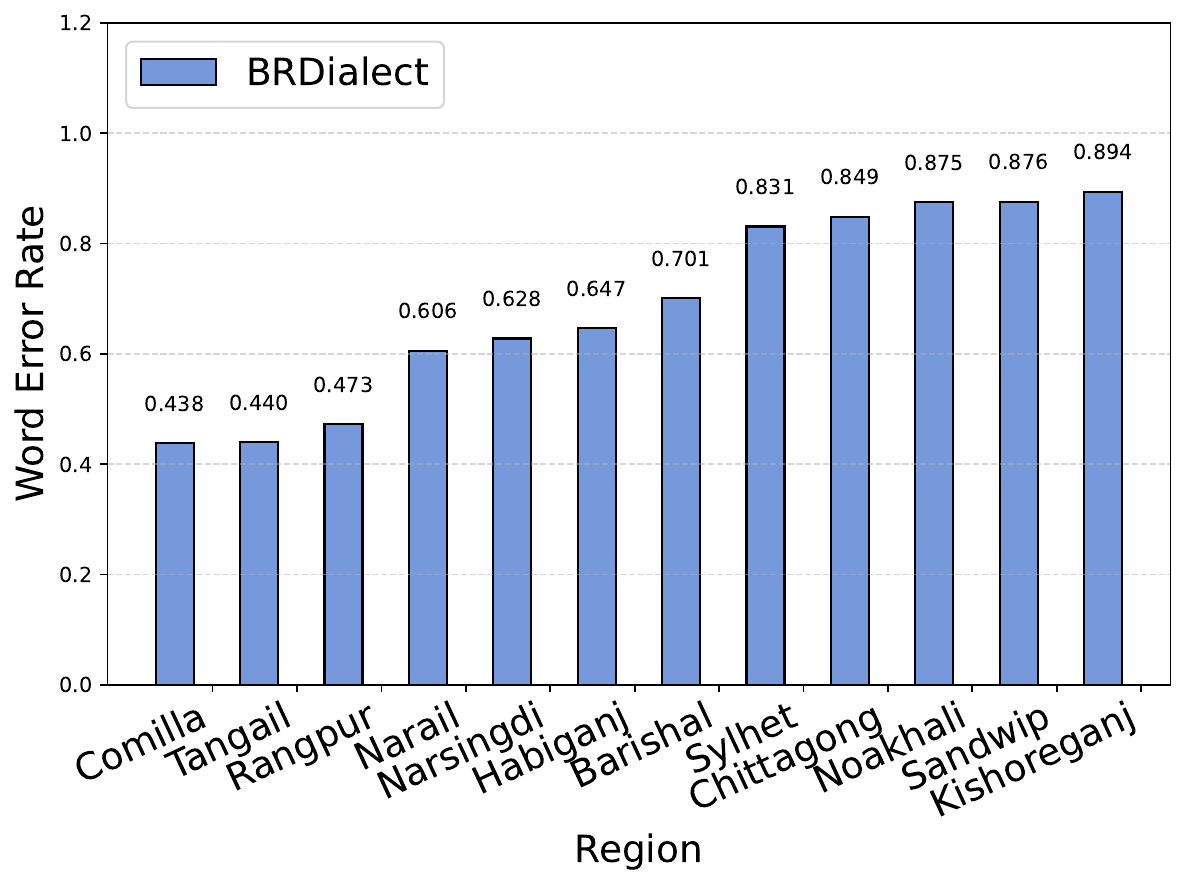}
    \caption{Regionwise word error rate distribution of the test set of the RegSpeech12 dataset. Transcriptions are generated using the BRDialect ASR system.}
    \label{fig:plot_wer}
\end{figure}

\paragraph{Performance across Regions}
We further evaluate BRDialect across individual regions. As shown in \cref{fig:plot_wer}, the ASR system performs well across most regions, with WER below 70\% in seven out of the twelve regions. The lowest WER, 0.438, is achieved for the Comilla region. Although the Comilla dialect is not included in the training data of BRDialect, its low WER demonstrates the model's strong generalization capability to unseen dialects.




\begin{table}[!t]
\centering

\begin{tabular}{@{}lcc@{}}
\toprule
\textbf{Processing} & \textbf{WER} $\downarrow$ & \textbf{CER} $\downarrow$\\
\midrule
{Noise Cancellation} & 0.876 & 0.497\\
\midrule
{No Noise Cancellation} & 0.865 & 0.452 \\
{5-gram KenLM Decoding} & 0.827 & 0.442 \\
{Unicode Normalization} & 0.796 & 0.420 \\
{Punctuation Removal} & \textbf{0.741} & \textbf{0.406} \\
\bottomrule
\end{tabular}
\caption{Impact of different types of processing on the performance of the BRDialect ASR system. Processing pipelines are combined from 
top to bottom consecutively for the group without noise cancellation.}
\label{tab:stt_processing}
\end{table}

\subsubsection{Ablation Study}
We analyze the effect of different preprocessing and postprocessing techniques on ASR performance. \cref{tab:stt_processing} summarizes the improvements observed with BRDialect. 

\paragraph{Impact of Noise Cancellation} 
We experiment with denoising the input audio before transcribing. The experimental results show that denoising with RNNoise slightly increases WER by 1.27\%. Aggressive denoising can remove speech cues necessary for the accurate transcription of regional dialects. For the remaining processing steps, we keep the original audio without denoising and combine processing pipelines. 

\paragraph{Impact of 5-gram KenLM Decoding} 
Integrating a 5-gram KenLM model during decoding improves WER from 0.865 to 0.827, a 4.39\% reduction. This confirms the value of training a KenLM language model with a Bengali regional text corpus for robust ASR performance \citep{rakib2023bangla}.

\paragraph{Impact of Unicode Normalization} 
Normalizing text further reduces WER to 0.796. This step is crucial since many Bengali characters can be represented in multiple ways, causing transcription inconsistencies. Normalizing text with the \texttt{BnUnicodeNormalizer} \citep{ansary2023unicode} ensures uniform representation, improving ASR performance.

\paragraph{Impact of Punctuation Removal} 
Since baseline ASR models do not generate punctuation, we remove punctuation from the RegSpeech12 transcriptions to ensure a fair comparison. Combined with previous steps, this yields the lowest WER of 0.741 and CER of 0.406. A detailed analysis of the processing configurations and their impact on the BRDialect ASR is provided in \cref{sec:app_processing_impact}.


\paragraph{Impact of High WER}
A Word Error Rate (WER) of {74.1}\% is relatively high for a general-purpose ASR system. However, dialectal speech recognition is inherently more challenging than the standard ASR task. The poor performance of the baseline models (\texttt{Whisper-medium-Bengali} and \texttt{IndicWav2Vec-Bengali}) validates this difficulty. In this context, BRDialect achieves a relative improvement of {12.41-33.98}\% over the baseline models, representing a substantial achievement. 

Within the BanglaTalk system, the integrated large language model (\texttt{GPT-4.1-nano}) effectively compensates for minor transcription errors. Leveraging its robust contextual understanding, the LLM can infer user intent even from noisy or imperfect ASR outputs, as illustrated in \cref{fig:interaction}. Most observed errors involve minor substitutions that do not significantly affect the intended meaning. 

Furthermore, explicitly prompting the LLM to interpret dialectal queries (\cref{fig:prompt}) enhances the system's ability to generate appropriate responses. The combination of a dialect-aware ASR model (BRDialect) and the powerful LLM (\texttt{GPT-4.1-nano}) ensures that the BanglaTalk system remains usable and effective despite relatively imperfect transcriptions.


\subsubsection{Levenshtein Distance Analysis}
\begin{figure}[!t]
    \centering
    \includegraphics[width=1\linewidth]{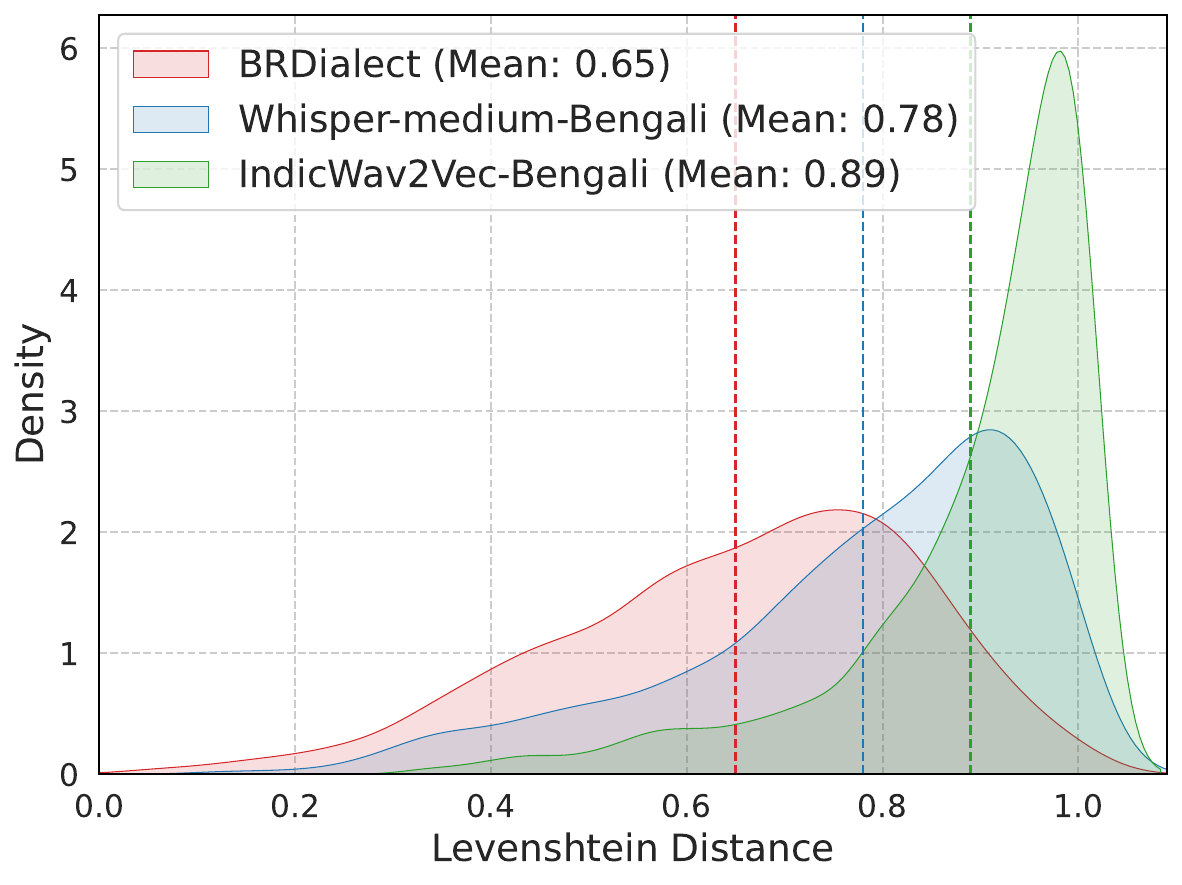}
    \caption{Distribution of Levenshtein distance for the best processing settings - without noise cancellation,  Bangla unicode normalization, and punctuation removal by three ASR systems on the RegSpeech12 dataset.}
    \label{fig:plot_levenshtein_distance}
\end{figure}

\cref{fig:plot_levenshtein_distance} illustrates the distribution of normalized Levenshtein distance between the ground-truth transcriptions and the outputs of the evaluated ASR systems under the best processing configuration -- No noise cancellation, Unicode normalization, and punctuation removal. For BRDialect, transcription quality is further refined during decoding using our trained 5-gram KenLM language model. A lower Levenshtein distance indicates higher transcription accuracy.

Among the models, BRDialect achieves the lowest mean distance of 0.65, demonstrating strong alignment with the reference transcriptions. \texttt{Whisper-medium-Bengali} performs moderately with a mean distance of 0.78, while \texttt{IndicWav2Vec-Bengali} consistently underperforms. The distribution of \texttt{IndicWav2Vec-Bengali} peaks sharply at 1, with the highest mean distance 0.89, highlighting substantial deviation from the ground truth.

In contrast, BRDialect exhibits a broader distribution, reflecting variability in performance -- some utterances are transcribed with high accuracy, while others are less. It achieves a measurable improvement of {16.67-26.97\%} compared to the baseline models, highlighting its effectiveness in handling dialectal diversity.


\begin{figure}[!t]
    \centering
    \includegraphics[width=1.0\linewidth]{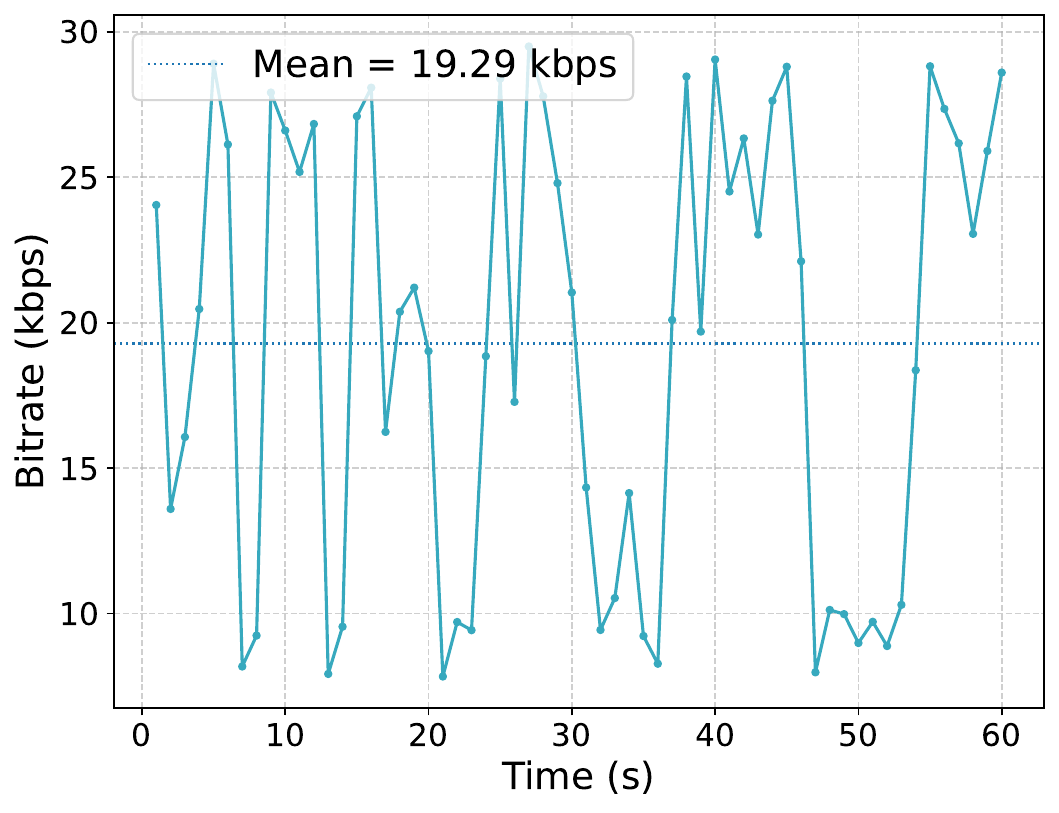}
    \caption{Uploading bitrate on the client side for a duration of one minute.}
    \label{fig:upload_bitrate}
\end{figure}

\subsection{Evaluating BanglaTalk}

\paragraph{Low Latency Audio Processing}

On the client side, each captured audio frame undergoes three sequential processing steps -- dynamic range compression, noise cancellation, and encoding with the Opus codec. The average processing time for an audio frame of duration 20 ms, is reported in \cref{tab:process_time}.

Among these, the encoding with the Opus codec is the most efficient, requiring only 0.56 ms per frame. This is consistent with the codec's design, which targets real-time, low-latency audio applications \citep{valin2016high}. Similarly, the DRC module introduces minimal computational overhead, averaging 1.31 ms per frame. In contrast, the noise cancellation with RNNoise introduces the highest processing time of 6.51 ms. 

RNNoise applies a lightweight neural network for speech enhancement, making it more resource-intensive than traditional signal processing methods. Additionally, RNNoise operates exclusively at a 48 kHz sample rate, while the BanglaTalk system relies on 16 kHz to ensure compatibility with ASR and VAD modules (see \cref{sec:capture_audio}). As a result, each frame must first be upsampled from 16 kHz to 48 kHz prior to noise cancellation and subsequently downsampled back to 16 kHz. These resampling operations introduce an additional delay of 1.40 ms for upsampling and 0.34 ms for downsampling. Compared to other ML-based noise cancellation systems \citep{cha2023dnoisenet}, RNNoise provides an excellent balance of speech quality and efficiency, making it highly suitable for real-time speech applications \citep{valin2018hybrid}.

Despite this, the total processing time (8.38 ms) remains well below the 20 ms frame duration, ensuring that all processing completes before the arrival of the next frame. This guarantees uninterrupted, real-time streaming without additional latency. 

\begin{table}[!t]
\centering
\begin{tabular}{@{}lr@{}}
\toprule
\textbf{Module} & \textbf{Process Time (ms)} \\
\midrule

\textbf{DRC}  & 1.31\\
\textbf{RNNoise}  & 6.51\\
\textbf{Encoding}  & \textbf{0.56}\\
\midrule
\textbf{Total} & 8.38\\
\bottomrule
\end{tabular}
\caption{Average processing times of a single audio frame of duration 20 ms.}
\label{tab:process_time}
\end{table}



\paragraph{Adaptive Bitrate Control}

In real-time assistive speech technologies, both end-to-end latency and bandwidth efficiency are crucial factors \citep{kaur2019intelligent}. Bandwidth efficiency is particularly important for underserved populations, such as those in rural areas of Bangladesh, where high-speed or expensive Internet connections are often unavailable.

To address this challenge, our system prioritizes bandwidth efficiency without compromising performance. Audio streams are encoded at a target bitrate of 24 kbps, significantly reducing bandwidth usage in the upload and download streams during client-server communication. \cref{fig:upload_bitrate} shows the upload bitrate of client-side for a one-minute audio stream.

We employ the variable bitrate (VBR) mode of the Opus codec, which dynamically adjusts the bitrate based on the characteristics of the audio signal -- allocating more bitrate to speech segments and conserving bitrate during silence. In the example shown in \cref{fig:upload_bitrate}, the average bitrate over one minute is only 19.29 kbps. This adaptive bitrate usage makes the system more accessible to users in economically disadvantaged regions \citep{osuagwu2013low}, who might otherwise be unable to benefit from assistive speech technologies.

Although applying RNNoise noise cancellation slightly increases the ASR system's Word Error Rate (WER) by {1.27}\% (\cref{tab:stt_processing}), it is retained in the client application due to its substantial benefit in reducing bandwidth usage. By removing background and foreground noise, the audio signal becomes more compressible. Without noise cancellation, the average upload bitrate of the audio (as shown in \cref{fig:upload_bitrate}) rises from {19.29} kbps to {23.6} kbps -- an {18.26}\% increase that is inefficient for client-side transmission. Given that the large language model (\texttt{GPT-4.1-nano}) demonstrates strong robustness in understanding minor transcription errors and generating accurate responses (\cref{fig:interaction}), the advantages of incorporating RNNoise into the client application outweigh its modest negative impact on ASR performance.





\paragraph{End-to-End Latency}
For real-time assistive speech systems, minimizing the delay between the end of a user's query and the beginning of the system's audio response is essential to ensure natural interactivity. The BanglaTalk system is designed with this principle in mind, introducing minimal overhead in both client and server-side processing.

We conduct rigorous testing and time analysis to measure the BanglaTalk system's end-to-end delay. As detailed in \cref{sec:processing_time_and_transcription}, the BRDialect ASR module of the BanglaTalk system introduces only a small delay, making it well-suited for real-time applications. The TTS systems evaluated in our study also exhibit low processing delay. \cref{sec:app_tts_eval} presents a detailed evaluation of TTS systems. Among them, the VITS-Bengali (male variant) is the best-performing model, achieving a high Mean Opinion Score (MOS) of 4.49 on the subset of the BanSpeech \citep{samin2024banspeech} dataset. 

To quantify the overall end-to-end delay of the BanglaTalk system, we simulate ten user queries in a conversation setting and measure the time elapsed from the end of the user queries to the start of the system's response. As shown in \cref{sec:end_to_end_delay}, the system achieves an average end-to-end delay of 4.9 seconds. This latency is acceptable for real-time assistive applications, ensuring smooth interactivity and an enhanced user experience.


\paragraph{Preliminary User Study}
To evaluate the real-world usability of the BanglaTalk system, we conduct a limited user study involving four native speakers of the Sylhet and Mymensingh dialects. The Sylhet dialect is selected due to its relative high WER ({0.831}), while the Mymensingh dialect is included to assess the generalizability of BRDialect to unseen dialects, as it is not part of the training set. Two native speakers from each region interact with BanglaTalk on general information and everyday task queries. Participants rate their interaction experience on a {1-5} scale, where {1} indicates a poor experience and {5} indicates an excellent experience. \cref{tab:rating} summarizes the results of five queries per user. Overall, BanglaTalk achieves a mean rating of {3.62}, indicating a generally positive user experience.

As illustrated in \cref{fig:interaction}, users from both regions interact successfully with the system despite minor transcription errors. Notably, users from the Mymensigh region, whose dialect is not included in BRDialect's training data, report a mean satisfaction rating of {3.73}, further demonstrating the model's strong generalization capability to unseen dialects. Participants also note that, despite their strong accent in regional dialects, BanglaTalk responds accurately and naturally, providing an interactive and effective speech-based experience.

\section{Related Work}

\subsection{Bengali Speech Assistants}
Several speech assistant systems have been developed for the Bengali language in recent years. The ALAPI system \citep{hasan2021alapi} introduces an open-domain Bengali conversational agent that processes recorded audio queries from users and generates response audio using AI techniques alongside a custom-built database. Shohojogi \citep{arnab2023shohojogi}, designed for the banking sector, provides voice-based customer support in Bengali. It integrates Wav2Vec2-based ASR \citep{baevski2020wav2vec}, query summarization, Google Text-to-Speech (gTTS), and a doc2vec model to retrieve relevant information for responses. Adrisya Sahayak \citep{sultan2021adrisya} presents a desktop-based virtual speech assistant for visually impaired Bengali speakers, supporting computer operations, peripheral devices, and home appliance control.

In the healthcare domain, a voice-enabled Artificial Conversational Entity has been developed to automate service interactions in Bengali \citep{pranto2021human}. This system leverages a domain-specific database and similarity-matching strategies to generate responses to user queries. Extending beyond monolingual assistants, Disha \citep{ullah2024disha} is a humanoid virtual assistant capable of interacting in both Bengali and English. It can address financial queries and perform real-time transactions. Beyond general-purpose applications, Bengali voice assistants have also been deployed in specialized domains such as providing information on metro rail services in Bangladesh \citep{rahman2024comprehensive}, assisting farmers with agricultural queries \citep{divakar2021farmer}, and offering accessible feminine healthcare support for marginalized women \citep{puja2024exploring}.

\subsection{English Speech Assistants}
Voice assistants such as Amazon Alexa, Apple Siri, Microsoft Cortana, and Google Assistant have become integral to modern human-computer interaction \citep{lopez2017alexa}. These systems, powered by artificial intelligence, are increasingly influencing daily life and societal practices \citep{subhash2020artificial}. Their development has been driven by advances in both signal processing and machine learning, which form the technological foundation of voice-based interaction \citep{haeb2019speech}.  

Beyond their technical design, researchers have examined the broader impact of these systems. For example, \citet{flavian2023effects} demonstrated that, compared to text-based recommendations, voice-based recommendations delivered by assistants have a stronger influence on consumer decision-making. User studies also indicate that voice assistants are most frequently employed for activities such as music, information search, and smart home (IoT) control \citep{ammari2019music}. Furthermore, their role has extended into education and healthcare contexts, where they support learning environments and assist older adults in managing everyday activities \citep{terzopoulos2020voice,oewel2023voice}.  


\subsection{Automatic Speech Recognition in Bengali}
Research on Automatic Speech Recognition (ASR) in Bengali has been addressed through both system development and survey-driven studies \citep{mridha2022study, tasnia2023overview, sultana2021recent}. The availability of large-scale datasets, such as Bengali Common Voice \citep{alam2022bengali}, OOD-Speech \citep{rakib2023ood}, and RegSpeech12 \citep{regspeech12}, has created significant opportunities for advancing Bengali ASR systems. Continuous Bengali ASR has been benchmarked on the SHRUTI corpus using DNN-HMM and GMM-HMM-based models \citep{al2019continuous}, achieving relatively low error rates. Leveraging CNN-RNN architectures, \citet{saha2021development} developed a gender- and speaker-independent ASR system. In Bangla-Wave \citep{rakib2023bangla}, the integration of an n-gram language model is shown to yield notable performance improvements. Furthermore, comparative analysis demonstrates that the Wav2Vec-BERT model outperforms Whisper on the Bengali Common Voice dataset \citep{ridoy2025adaptability}.


\section{Conclusion \& Future Work}

In this work, we present BanglaTalk, the first real-time, end-to-end conversational speech assistant designed specifically for Bengali regional dialects. The system integrates both client and server applications, with low-latency audio signal processing implemented on both ends. To ensure efficient network transmission, speech data is transmitted at a low bandwidth of 24 kbps, making the system accessible to a broad range of users. Our developed BRDialect ASR system, integrated into the BanglaTalk pipeline, effectively transcribes Bengali regional dialects, achieving an overall word error rate of 74.1\% and a character error rate of 40.6\% on the RegSpeech12 dataset, which spans 12 regions of Bangladesh. For speech synthesis, the VITS-Bengali TTS model (male) incorporated into the system attains a mean opinion score (MOS) of 4.49 on a subset of the BanSpeech dataset, enhancing the naturalness and human-like quality of the generated voice. Furthermore, the system maintains a low end-to-end delay of 4.9 seconds, ensuring a highly interactive user experience. These performance metrics demonstrate the effectiveness of the system for real-time communication.

\section*{Limitations}



The BRDialect ASR system is trained on regional speech data covering ten regions of Bangladesh. 
Regions not included in the training data may experience less accurate transcriptions when using the BanglaTalk system. 
Incorporating speech data from the remaining regions of Bangladesh is expected to significantly improve the performance of the BRDialect ASR system. Although BanglaTalk reduces end-to-end delay and is highly interactive, several limitations remain:


\begin{itemize}
    \item User interruption: In the current system, the assistant continues speaking until the utterance is completed. Adding the capability to handle user interruptions would make conversations more natural and interactive.
    \item Speaker verification: The system does not verify the speaker. If another person speaks during the user’s conversation, their speech is transcribed, which negatively affects performance. Incorporating speaker verification would mitigate this issue.
    \item Concurrent conversations: At present, only one conversation can be executed at a time. Supporting multiple concurrent conversations would increase system availability and usability.
    \item User study coverage: Feedback has so far been collected from a limited set of regional speakers. A broader user study covering all regions of Bangladesh would provide deeper insights into system performance, user acceptance, and overall impact.
\end{itemize}

\bibliography{custom}

@inproceedings{al2019continuous,
  title={Continuous bengali speech recognition based on deep neural network},
  author={Al Amin, Md Alif and Islam, Md Towhidul and Kibria, Shafkat and Rahman, Mohammad Shahidur},
  booktitle={2019 international conference on electrical, computer and communication engineering (ECCE)},
  pages={1--6},
  year={2019},
  organization={IEEE}
}

@article{samin2024banspeech,
  title={BanSpeech: A Multi-Domain Bangla Speech Recognition Benchmark Toward Robust Performance in Challenging Conditions},
  author={Samin, Ahnaf Mozib and Kobir, M Humayon and Rafee, Md Mushtaq Shahriyar and Ahmed, M Firoz and Hasan, Mehedi and Ghosh, Partha and Kibria, Shafkat and Rahman, M Shahidur},
  journal={IEEE Access},
  volume={12},
  pages={34527--34538},
  year={2024},
  publisher={IEEE}
}

@article{giannoulis2012digital,
  title={Digital dynamic range compressor design—A tutorial and analysis},
  author={Giannoulis, Dimitrios and Massberg, Michael and Reiss, Joshua D},
  journal={Journal of the Audio Engineering Society},
  volume={60},
  number={6},
  pages={399--408},
  year={2012},
  publisher={Audio Engineering Society}
}

@inproceedings{valin2018hybrid,
  title={A hybrid DSP/deep learning approach to real-time full-band speech enhancement},
  author={Valin, Jean-Marc},
  booktitle={2018 IEEE 20th international workshop on multimedia signal processing (MMSP)},
  pages={1--5},
  year={2018},
  organization={IEEE}
}

@article{valin2016high,
  title={High-quality, low-delay music coding in the opus codec},
  author={Valin, Jean-Marc and Maxwell, Gregory and Terriberry, Timothy B and Vos, Koen},
  journal={arXiv preprint arXiv:1602.04845},
  year={2016}
}

@misc{audio1996rfc1889,
  title={RFC1889: RTP: A transport protocol for real-time applications},
  author={Audio-Video Transport Working Group and Schulzrinne, H and Casner, S and Frederick, R and Jacobson, V and others},
  year={1996},
  publisher={RFC Editor}
}

@techreport{postel1980user,
  title={User datagram protocol},
  author={Postel, Jon},
  year={1980}
}

@misc{SileroVAD,
  author = {Silero Team},
  title = {Silero VAD: pre-trained enterprise-grade Voice Activity Detector (VAD), Number Detector and Language Classifier},
  year = {2024},
  publisher = {GitHub},
  journal = {GitHub repository},
  howpublished = {\url{https://github.com/snakers4/silero-vad}},
  commit = {insert_some_commit_here},
  email = {hello@silero.ai}
}

@article{cha2023dnoisenet,
  title={DNoiseNet: Deep learning-based feedback active noise control in various noisy environments},
  author={Cha, Young-Jin and Mostafavi, Alireza and Benipal, Sukhpreet S},
  journal={Engineering Applications of Artificial Intelligence},
  volume={121},
  pages={105971},
  year={2023},
  publisher={Elsevier}
}

@article{ridoy2025adaptability,
  title={Adaptability of ASR Models on Low-Resource Language: A Comparative Study of Whisper and Wav2Vec-BERT on Bangla},
  author={Ridoy, Md Sazzadul Islam and Akter, Sumi and Rahman, Md Aminur},
  journal={arXiv preprint arXiv:2507.01931},
  year={2025}
}

@article{rakib2023ood,
  title={Ood-speech: A large bengali speech recognition dataset for out-of-distribution benchmarking},
  author={Rakib, Fazle Rabbi and Dip, Souhardya Saha and Alam, Samiul and Tasnim, Nazia and Shihab, Md Istiak Hossain and Ansary, Md Nazmuddoha and Hossen, Syed Mobassir and Meghla, Marsia Haque and Mamun, Mamunur and Sadeque, Farig and others},
  journal={arXiv preprint arXiv:2305.09688},
  year={2023}
}

@article{alam2022bengali,
  title={Bengali common voice speech dataset for automatic speech recognition},
  author={Alam, Samiul and Sushmit, Asif and Abdullah, Zaowad and Nakkhatra, Shahrin and Ansary, MD and Hossen, Syed Mobassir and Mehnaz, Sazia Morshed and Reasat, Tahsin and Humayun, Ahmed Imtiaz},
  journal={arXiv preprint arXiv:2206.14053},
  year={2022}
}

@article{sultana2021recent,
  title={Recent advancement in speech recognition for bangla: A survey},
  author={Sultana, Sadia and Rahman, M Shahidur and Iqbal, M Zafar},
  journal={International Journal of Advanced Computer Science and Applications},
  volume={12},
  number={3},
  year={2021},
  publisher={Science and Information (SAI) Organization Limited}
}

@inproceedings{tasnia2023overview,
  title={An overview of bengali speech recognition: Methods, challenges, and future direction},
  author={Tasnia, Nabila and Islam, Mahidul and Rony, Mahi Shahriar and Tanzim, Nishat and Hasib, Khan Md and Alam, Mohammad Shafiul},
  booktitle={2023 IEEE 13th Annual Computing and Communication Workshop and Conference (CCWC)},
  pages={0873--0878},
  year={2023},
  organization={IEEE}
}

@article{mridha2022study,
  title={A study on the challenges and opportunities of speech recognition for Bengali language},
  author={Mridha, Muhammad Firoz and Ohi, Abu Quwsar and Hamid, Md Abdul and Monowar, Muhammad Mostafa},
  journal={Artificial Intelligence Review},
  volume={55},
  number={4},
  pages={3431--3455},
  year={2022},
  publisher={Springer}
}

@inproceedings{rakib2023bangla,
  title={Bangla-wave: Improving bangla automatic speech recognition utilizing n-gram language models},
  author={Rakib, Mohammed and Hossain, Md Ismail and Mohammed, Nabeel and Rahman, Fuad},
  booktitle={Proceedings of the 2023 12th International Conference on Software and Computer Applications},
  pages={297--301},
  year={2023}
}

@inproceedings{saha2021development,
  title={Development of a bangla speech to text conversion system using deep learning},
  author={Saha, Srijoni and others},
  booktitle={2021 Joint 10th International Conference on Informatics, Electronics \& Vision (ICIEV) and 2021 5th International Conference on Imaging, Vision \& Pattern Recognition (icIVPR)},
  pages={1--7},
  year={2021},
  organization={IEEE}
}

@article{baevski2020wav2vec,
  title={wav2vec 2.0: A framework for self-supervised learning of speech representations},
  author={Baevski, Alexei and Zhou, Yuhao and Mohamed, Abdelrahman and Auli, Michael},
  journal={Advances in neural information processing systems},
  volume={33},
  pages={12449--12460},
  year={2020}
}

@misc{hasan_saim_mostafa_2024_bengaliASR_indicwav2vec,
  author       = {Hasan, Jakir and Saim, Md. Ataullha and Mostafa, Radeen},
  title        = {Improving Bengali ASR for Regional Dialects with IndicWav2Vec: A Competition Approach},
  year         = {2024},
  howpublished = {Preprint, ResearchGate},
  doi          = {10.13140/RG.2.2.10463.27049},
  url          = {https://www.researchgate.net/publication/381922998_Improving_Bengali_ASR_for_Regional_Dialects_with_IndicWav2Vec_A_Competition_Approach}
}

@misc{ben10,
  author       = {Ahmed Imtiaz Humayun and farigys and Mohaymen Ul Anam and Rubayet Sabbir Faruque and S. M. Jishanul Islam and Sushmit and Tahsin},
  title        = {ASR for Regional Dialects},
  year         = {2024},
  howpublished = {Kaggle competition},
  note         = {Accessed: 2025-09-19},
  url          = {https://kaggle.com/competitions/ben10}
}

@inproceedings{javed2022towards,
  title={Towards building asr systems for the next billion users},
  author={Javed, Tahir and Doddapaneni, Sumanth and Raman, Abhigyan and Bhogale, Kaushal Santosh and Ramesh, Gowtham and Kunchukuttan, Anoop and Kumar, Pratyush and Khapra, Mitesh M},
  booktitle={Proceedings of the aaai conference on artificial intelligence},
  volume={36},
  number={10},
  pages={10813--10821},
  year={2022}
}

@inproceedings{heafield2011kenlm,
  title={KenLM: Faster and smaller language model queries},
  author={Heafield, Kenneth},
  booktitle={Proceedings of the sixth workshop on statistical machine translation},
  pages={187--197},
  year={2011}
}

@inproceedings{liang2023dynamic,
  title={Dynamic speech endpoint detection with regression targets},
  author={Liang, Dawei and Su, Hang and Singh, Tarun and Mahadeokar, Jay and Puri, Shanil and Zhu, Jiedan and Thomaz, Edison and Seltzer, Mike},
  booktitle={ICASSP 2023-2023 IEEE International Conference on Acoustics, Speech and Signal Processing (ICASSP)},
  pages={1--5},
  year={2023},
  organization={IEEE}
}

@article{dam2024complete,
  title={A complete survey on llm-based ai chatbots},
  author={Dam, Sumit Kumar and Hong, Choong Seon and Qiao, Yu and Zhang, Chaoning},
  journal={arXiv preprint arXiv:2406.16937},
  year={2024}
}

@inproceedings{raju2019bangla,
  title={A bangla text-to-speech system using deep neural networks},
  author={Raju, Rajan Saha and Bhattacharjee, Prithwiraj and Ahmad, Arif and Rahman, Mohammad Shahidur},
  booktitle={2019 International Conference on Bangla Speech and Language Processing (ICBSLP)},
  pages={1--5},
  year={2019},
  organization={IEEE}
}

@article{pratap2024scaling,
  title={Scaling speech technology to 1,000+ languages},
  author={Pratap, Vineel and Tjandra, Andros and Shi, Bowen and Tomasello, Paden and Babu, Arun and Kundu, Sayani and Elkahky, Ali and Ni, Zhaoheng and Vyas, Apoorv and Fazel-Zarandi, Maryam and others},
  journal={Journal of Machine Learning Research},
  volume={25},
  number={97},
  pages={1--52},
  year={2024}
}

@incollection{kaur2019intelligent,
  title={Intelligent voice bots for digital banking},
  author={Kaur, Ravneet and Sandhu, Ravtej Singh and Gera, Ayush and Kaur, Tarlochan and Gera, Purva},
  booktitle={Smart Systems and IoT: Innovations in Computing: Proceeding of SSIC 2019},
  pages={401--408},
  year={2019},
  publisher={Springer}
}

@misc{regspeech12,
  author       = {Md. Rezuwan Hassan and Azmol Hossain and Kanij Fatema and Rubayet Sabbir Faruque and Tanmoy Shome and Ruwad Naswan and Trina Chakraborty and Tawsif Tashwar Dipto and Md Foriduzzaman Zihad and Nazmuddoha Ansary and Asif Sushmit and Ahmed Imtiaz Humayun and Tahsin Reasat and Md Mehedi Hasan Shawon and Md. Golam Rabiul Alam and Farig Sadeque},
  title        = {RegSpeech12: A Regional Corpus of Bengali Spontaneous Speech Across Dialects},
  howpublished = {Kaggle Dataset},
  year         = {2025},       
  note         = {Accessed: 2025-09-19},  
  url          = {https://www.kaggle.com/datasets/mdrezuwanhassan/regspeech12}
}

@article{ansary2023unicode,
  title={Unicode Normalization and Grapheme Parsing of Indic Languages},
  author={Ansary, Nazmuddoha and Adib, Quazi Adibur Rahman and Reasat, Tahsin and Sushmit, Asif Shahriyar and Humayun, Ahmed Imtiaz and Mehnaz, Sazia and Fatema, Kanij and Rashid, Mohammad Mamun Or and Sadeque, Farig},
  journal={arXiv preprint arXiv:2306.01743},
  year={2023}
}

@inproceedings{oewel2023voice,
  title={Voice assistant use in long-term care},
  author={Oewel, Bruna and Ammari, Tawfiq and Brewer, Robin N},
  booktitle={Proceedings of the 5th International Conference on Conversational User Interfaces},
  pages={1--10},
  year={2023}
}

@article{terzopoulos2020voice,
  title={Voice assistants and smart speakers in everyday life and in education},
  author={Terzopoulos, George and Satratzemi, Maya},
  journal={Informatics in Education},
  volume={19},
  number={3},
  pages={473--490},
  year={2020},
  publisher={Vilnius University Institute of Data Science and Digital Technologies}
}

@article{ammari2019music,
  title={Music, search, and IoT: How people (really) use voice assistants},
  author={Ammari, Tawfiq and Kaye, Jofish and Tsai, Janice Y and Bentley, Frank},
  journal={ACM Transactions on Computer-Human Interaction (TOCHI)},
  volume={26},
  number={3},
  pages={1--28},
  year={2019},
  publisher={ACM New York, NY, USA}
}

@article{haeb2019speech,
  title={Speech processing for digital home assistants: Combining signal processing with deep-learning techniques},
  author={Haeb-Umbach, Reinhold and Watanabe, Shinji and Nakatani, Tomohiro and Bacchiani, Michiel and Hoffmeister, Bjorn and Seltzer, Michael L and Zen, Heiga and Souden, Mehrez},
  journal={IEEE Signal processing magazine},
  volume={36},
  number={6},
  pages={111--124},
  year={2019},
  publisher={IEEE}
}

@article{flavian2023effects,
  title={Effects of voice assistant recommendations on consumer behavior},
  author={Flavi{\'a}n, Carlos and Akdim, Khaoula and Casal{\'o}, Luis V},
  journal={Psychology \& Marketing},
  volume={40},
  number={2},
  pages={328--346},
  year={2023},
  publisher={Wiley Online Library}
}

@inproceedings{subhash2020artificial,
  title={Artificial intelligence-based voice assistant},
  author={Subhash, S and Srivatsa, Prajwal N and Siddesh, S and Ullas, A and Santhosh, B},
  booktitle={2020 Fourth world conference on smart trends in systems, security and sustainability (WorldS4)},
  pages={593--596},
  year={2020},
  organization={IEEE}
}

@inproceedings{lopez2017alexa,
  title={Alexa vs. Siri vs. Cortana vs. Google assistant: a comparison of speech-based natural user interfaces},
  author={L{\'o}pez, Gustavo and Quesada, Luis and Guerrero, Luis A},
  booktitle={International conference on applied human factors and ergonomics},
  pages={241--250},
  year={2017},
  organization={Springer}
}

@misc{BengaliAIASRSubmission2023,
  author       = {Tugstugi},
  title        = {Bengali AI ASR Submission},
  year         = {2023},
  howpublished = {Kaggle dataset},
  note         = {Accessed: 2025-09-19},
  url          = {https://www.kaggle.com/datasets/tugstugi/bengali-ai-asr-submission}
}

@inproceedings{kim2021conditional,
  title={Conditional variational autoencoder with adversarial learning for end-to-end text-to-speech},
  author={Kim, Jaehyeon and Kong, Jungil and Son, Juhee},
  booktitle={International Conference on Machine Learning},
  pages={5530--5540},
  year={2021},
  organization={PMLR}
}

@misc{ComprehensiveBanglaTTS2023,
  author       = {Mobassir Hossen},
  title        = {Comprehensive Bangla TTS},
  year         = {2023},
  howpublished = {Kaggle dataset},
  note         = {Accessed: 2025-09-19},
  url          = {https://www.kaggle.com/datasets/mobassir/comprehensive-bangla-tts}
}

@article{streijl2016mean,
  title={Mean opinion score (MOS) revisited: methods and applications, limitations and alternatives},
  author={Streijl, Robert C and Winkler, Stefan and Hands, David S},
  journal={Multimedia Systems},
  volume={22},
  number={2},
  pages={213--227},
  year={2016},
  publisher={Springer}
}

@inproceedings{dobson2000developments,
  title={Developments in audio file formats},
  author={Dobson, Richard W},
  booktitle={ICMC},
  year={2000}
}

@article{hoy2018alexa,
  title={Alexa, Siri, Cortana, and more: an introduction to voice assistants},
  author={Hoy, Matthew B},
  journal={Medical reference services quarterly},
  volume={37},
  number={1},
  pages={81--88},
  year={2018},
  publisher={Taylor \& Francis}
}

@article{xu2022research,
  title={Research on interpolation and data fitting: Basis and applications},
  author={Xu, Yijie and Xu, Runqi},
  journal={arXiv preprint arXiv:2208.11825},
  year={2022}
}

@article{sundvall2014opus,
  title={Opus audio codec in mobile networks},
  author={Sundvall, Mika},
  year={2014}
}

@article{dutsinma2022systematic,
  title={A systematic review of voice assistant usability: An ISO 9241--11 approach},
  author={Dutsinma, Faruk Lawal Ibrahim and Pal, Debajyoti and Funilkul, Suree and Chan, Jonathan H},
  journal={SN computer science},
  volume={3},
  number={4},
  pages={267},
  year={2022},
  publisher={Springer}
}

@article{osuagwu2013low,
  title={Low and Expensive Bandwidth Remains Key Bottleneck for Nigeria’s Internet Diffusion: A Proposal for a Solution Model.},
  author={Osuagwu, OE and Okide, S and Edebatu, D and Udoka, E},
  journal={West African Journal of Industrial and Academic Research},
  volume={7},
  number={1},
  pages={14--30},
  year={2013}
}

@article{islam2025banglalem,
  title={BanglaLem: a transformer-based bangla lemmatizer with an enhanced dataset},
  author={Islam, Md Fuadul and Hasan, Jakir and Islam, Md Ashikul and Dewan, Prato and Rahman, M Shahidur},
  journal={Systems and Soft Computing},
  pages={200244},
  year={2025},
  publisher={Elsevier}
}

@inproceedings{hasan2024credibility,
  title={Credibility Analysis of Robot Speech Based on Bangla Language Dialect},
  author={Hasan, Md Nahid and Azim, Raiyan and Sharmin, Sadia},
  booktitle={2024 IEEE International Conference on Computing, Applications and Systems (COMPAS)},
  pages={1--6},
  year={2024},
  organization={IEEE}
}

@techreport{schulzrinne2003rtp,
  title={RTP: A transport protocol for real-time applications},
  author={Schulzrinne, Henning and Casner, Stephen and Frederick, Ron and Jacobson, Van},
  year={2003}
}

@article{puja2024exploring,
  title={Exploring the barriers to feminine healthcare access among marginalized women in Bangladesh and facilitating access through a voice bot},
  author={Puja, Sreya Sanyal and Neha, Nahian Noor and Alif, Ofia Rahman and Sultan, Tarannaum Jahan and Husna, Md Golam Zel Asmaul and Jahan, Ishrat and Noor, Jannatun},
  journal={Heliyon},
  volume={10},
  number={14},
  year={2024},
  publisher={Elsevier}
}

@inproceedings{divakar2021farmer,
  title={FARMER’S assistant using ai voice bot},
  author={Divakar, MS and Kumar, Vimal and DE, Martina Jaincy and Kalpana, RA and RM, Sanjai Kumar and others},
  booktitle={2021 3rd International Conference on Signal Processing and Communication (ICPSC)},
  pages={527--531},
  year={2021},
  organization={IEEE}
}

@phdthesis{rahman2024comprehensive,
  title={A comprehensive NLP-based voice assistant system for streamlined information retrieval in metro rail services of Bangladesh},
  author={Rahman, MD and Alamgir, Adnan and Chowdhury, Shaheedul Haque and Mushtari, Maliha and Anzum, Wasim and others},
  year={2024},
  school={Brac University}
}

@inproceedings{ullah2024disha,
  title={Disha: A Bilingual Humanoid Virtual Assistant},
  author={Ullah, Md Rifat and Mahbub, Md Nafees and Hakim, Md Azizul and Sultana, Yeasmin and Alam, Iftakhar},
  booktitle={2024 6th International Conference on Electrical Engineering and Information \& Communication Technology (ICEEICT)},
  pages={457--462},
  year={2024},
  organization={IEEE}
}

@inproceedings{hasan2021alapi,
  title={ALAPI: An automated voice chat system in Bangla language},
  author={Hasan, Md Mehedi and Roy, Auronno and Hasan, Md Tariq},
  booktitle={2021 International Conference on Electronics, Communications and Information Technology (ICECIT)},
  pages={1--4},
  year={2021},
  organization={IEEE}
}

@phdthesis{arnab2023shohojogi,
  title={Shohojogi: An automated voice chat system in the bangla language for the banking system},
  author={Arnab, Kabir Abdur Rahman and Hossain, Ashiqual and Nabi, Istihad and Hossain, Muhammad Iqbal},
  year={2023},
  school={Ph. D. Dissertation. https://doi. org/10. 13140/RG. 2.2. 34757.22240/1}
}

@inproceedings{sultan2021adrisya,
  title={Adrisya sahayak: A bangla virtual assistant for visually impaired},
  author={Sultan, Md Rakibuz and Hoque, Md Moinul and Heeya, Farah Ulfath and Ahmed, Iftiquar and Ferdouse, Md Redwanul and Mubin, Shikder Mejbah Ahmed},
  booktitle={2021 2nd international conference on robotics, electrical and signal processing techniques (ICREST)},
  pages={597--602},
  year={2021},
  organization={IEEE}
}

@inproceedings{pranto2021human,
  title={Human-robot interaction in bengali language for healthcare automation integrated with speaker recognition and artificial conversational entity},
  author={Pranto, Shehan Irteza and Nabid, Rahad Arman and Samin, Ahnaf Mozib and Mohammed, Nabeel and Sarker, Farhana and Huda, Mohammad Nurul and Mamun, Khondaker A},
  booktitle={2021 3rd International Conference on Electrical \& Electronic Engineering (ICEEE)},
  pages={13--16},
  year={2021},
  organization={IEEE}
}

\clearpage
\appendix

\section{Algorithms}


\subsection{Upsampling}
The upsampling algorithm described in \cref{alg:upsample} increases the audio sample rate by generating intermediate samples through linear interpolation.


\begin{algorithm}[htbp]
\caption{Upsampling with linear interpolation increases the sample rate by inserting new sample values between consecutive samples. The intermediate values are calculated using the previous-current sample pair and the slope between them.}
\label{alg:upsample}
\begin{algorithmic}[1]
\Require Input frame $x \in \mathbb{Z}^{N}$, 
         ratio $r$, 
         previous value $p$
\Ensure  Upsampled array $y \in \mathbb{Z}^{N \cdot r}$
\State $y \gets$ array of zeros with length $N \cdot r$
\State $k \gets 1$
\For{$i \gets 1$ to $N$}
    \State $cur \gets x[i]$
    \State $\Delta \gets (cur - p)/r$ 
    \For{$j \gets 0$ to $r-1$}
        \State $v \gets p + j \cdot \Delta$
        \State Clip $v$ into range $[-32768, 32767]$
        \State $y[k] \gets \lfloor v \rfloor$; $k \gets k+1$
    \EndFor
    \State $p \gets cur$
\EndFor
\State \Return $y$
\end{algorithmic}
\end{algorithm}

\subsection{Downsampling}
The downsampling algorithm described in \cref{alg:downsample} reduces the audio sample rate by discarding intermediate samples.

\begin{algorithm}[!h]
\caption{Downsampling by an integer factor $r$ reduces the sample rate by keeping every $r-th$ sample from the speech signal.}
\label{alg:downsample}
\begin{algorithmic}[1]
\Require Input frame $x \in \mathbb{Z}^{L}$, ratio $r \in \mathbb{N}, r \ge 1$
\Ensure  Downsampled array $y \in \mathbb{Z}^{\lfloor L/r \rfloor}$
\State $N \gets \left\lfloor \dfrac{L}{r} \right\rfloor$
\State $y \gets$ array of zeros with length $N$
\State $k \gets 1$
\For{$i \gets 1$ to $L$ \textbf{step} $r$}
    \State $y[k] \gets x[i]$; \hspace{0.7em} $k \gets k+1$
    \If{$k > N$} \textbf{break} \EndIf
\EndFor
\State \Return $y$
\end{algorithmic}
\end{algorithm}

 \label{sec:app_algo}

\section{The RTP Packet Structure}
\label{sec:app_rtp_structure}
\begin{figure*}[!t]
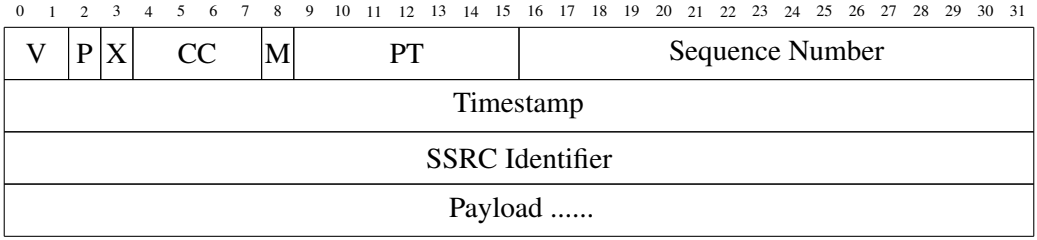

\centering
\begin{bytefield}[bitwidth=1.1em]{32}
    \bitheader{0-31}\\
    \bitbox{2}{V} & \bitbox{1}{P} & \bitbox{1}{X} & \bitbox{4}{CC} & \bitbox{1}{M} & \bitbox{7}{PT} & \bitbox{16}{Sequence Number}\\
    \bitbox{32}{Timestamp}\\
    \bitbox{32}{SSRC Identifier}\\
    \bitbox{32}{Payload ......}\\
\end{bytefield}
\caption{RTP packet structure with a 12-byte header followed by the audio payload.}
\label{fig:rtp_packet}
\end{figure*}

The structure of a standard RTP packet, which comprises a 12-byte header, is shown in \cref{fig:rtp_packet}.

 \label{sec:app_rtp}

\section{Dataset}
\label{sec:app_dataset}

\subsection{Ben10 Dataset Analysis}
\label{sec:app_ben10}


The Ben10 dataset \citep{ben10} comprises speech recordings from 373 speakers across ten distinct regions of Bangladesh. The training set contains over 63 hours of audio sampled at 16 kHz. \cref{tab:tab_ben10} presents the region-wise distribution of audio files within the training set. Since the transcripts of the Ben10 test set are not publicly available, we instead employ the test set of the RegSpeech12 dataset \citep{regspeech12} for evaluation in our study.


\begin{table}[htbp]
\centering

\begin{tabular}{@{}lc@{}}
\toprule
\multicolumn{2}{c}{\textbf{Training Set}} \\

\cmidrule{1-2}
\textbf{Region} & \textbf{Audio File Count}\\

\cmidrule{1-2}

Barishal & 796 \\
Chittagong & 1406 \\
Habiganj & 940 \\
Kishoreganj & 1638 \\
Narail & 1488 \\
Narsingdi & 1098 \\
Rangpur & 1037 \\
Sandwip & 1049 \\
Sylhet & 2903 \\
Tangail & 987 \\

\cmidrule{1-2}
\textbf{Total} & \textbf{13342} \\

\bottomrule

\end{tabular}
\caption{Training set distribution of the Ben10 dataset}
\label{tab:tab_ben10}
\end{table}

\subsection{RegSpeech12 Dataset Analysis}
\label{sec:app_regspeech12}


The RegSpeech12 dataset \citep{regspeech12} is a spontaneous speech corpus encompassing twelve regional dialects of Bangladesh, comprising approximately 100 hours of speech data. In this study, we utilize the test split of the dataset, which contains around 10 hours of recordings. \cref{tab:tab_regspeech12} presents the region-wise distribution of the test set.



\begin{table}[!h]
\centering

\begin{tabular}{@{}lc@{}}
\toprule
\multicolumn{2}{c}{\textbf{Test Set}} \\

\cmidrule{1-2}
\textbf{Region} & \textbf{Audio File Count}\\

\cmidrule{1-2}

Barishal & 101 \\
Chittagong & 176 \\
Comilla & 32 \\
Habiganj & 117 \\
Kishoreganj & 205 \\
Narail & 186 \\
Narsingdi & 137 \\
Noakhali & 28 \\
Rangpur & 130 \\
Sandwip & 131 \\
Sylhet & 762 \\
Tangail & 127 \\

\cmidrule{1-2}
\textbf{Total} & \textbf{2132} \\

\bottomrule

\end{tabular}
\caption{Test set distribution of the RegSpeech12 dataset}
\label{tab:tab_regspeech12}
\end{table}

\section{Prompt to LLM}
\label{sec:llm_prompt}


To generate responses to spoken user queries transcribed by the dialect-aware ASR system, we provide the prompt illustrated in \cref{fig:prompt} to the GPT-4.1-nano model.

\begin{center}
\begin{minipage}{\columnwidth} 
\begin{tcolorbox}[colback=gray!5!white,
                  colframe=gray!50!black,
                  colbacktitle=gray!75!black]

   \textbf{System Prompt: } You are a helpful chatbot who understands Bengali regional dialects and only speaks standard Bengali. Please be concise and end every sentence with $\texttt{\{|\}}$.\\[6pt]
   \textbf{User Prompt: } Please generate a response for only the valid query. For an invalid query, print only a $\texttt{\{\$\}}$. Here is the query in the Bengali regional dialect $\texttt{\{user\_query\}}$.

\end{tcolorbox}
\captionof{figure}{Prompt for LLM to generate response for query in Bengali regional dialect.}
\label{fig:prompt}
\end{minipage}
\end{center} \label{sec:app_prompt}

\section{ASR System Evaluation}
\label{sec:result_appendix}

\subsection{ASR Evaluation Metrics}
\label{sec:wer_cer_formula}

Word Error Rate (WER) and Character Error Rate (CER) are the standard metrics for evaluating the performances of ASR systems.
The formula for Word Error Rate is:
\begin{equation}
\text{WER} = \frac{S + D + I}{N} \times 100\%,
\end{equation}
where 
$S$ is the number of substitutions, 
$D$ is the number of deletions, 
$I$ is the number of insertions, 
and $N$ is the total number of words in the reference transcription.

Similarly, the formula for Character Error Rate is:
\begin{equation}
\text{CER} = \frac{S_c + D_c + I_c}{N_c} \times 100\%,
\end{equation}
where 
$S_c$ is the number of substitutions, 
$D_c$ is the number of deletions, and 
$I_c$ is the number of insertions at the character level.
$N_c$ is the total number of characters in the reference transcription.


\subsection{Impact of Processing on BRDialect ASR System}
\label{sec:app_processing_impact}


Processing pipelines play a critical role in shaping the performance of ASR systems. \cref{tab:tab_impact_brdialect} presents the impact of 5-gram KenLM decoding, noise cancellation, text normalization, and punctuation removal on the BRDialect ASR system.

\begin{table}[!t]
\centering
\begin{tabular}{@{}cccccc@{}}
\toprule
\textbf{KD} & \textbf{NC} & \textbf{UN} & \textbf{PR} & \textbf{WER} & \textbf{CER}\\

\cmidrule{1-6}

\ding{51} & \ding{53} & \ding{53} & \ding{53} & 0.827 & 0.442\\ 
\ding{51} & \ding{53} & \ding{53} & \ding{51} & 0.796 & 0.420\\
\ding{51} & \ding{53} & \ding{51} & \ding{53} & 0.781 & 0.430\\
\ding{51} & \ding{53} & \ding{51} & \ding{51} & \textbf{0.741} & \textbf{0.406}\\
\ding{51} & \ding{51} & \ding{53} & \ding{53} & 0.838 & 0.493\\ 
\ding{51} & \ding{51} & \ding{53} & \ding{51} & 0.813 & 0.474\\
\ding{51} & \ding{51} & \ding{51} & \ding{53} & 0.801 & 0.479\\
\ding{51} & \ding{51} & \ding{51} & \ding{51} & 0.770 & 0.458\\

\ding{53} & \ding{53} & \ding{53} & \ding{53} & 0.865 & 0.452\\ 
\ding{53} & \ding{53} & \ding{53} & \ding{51} & 0.834 & 0.444\\
\ding{53} & \ding{53} & \ding{51} & \ding{53} & 0.834 & 0.429\\
\ding{53} & \ding{53} & \ding{51} & \ding{51} & 0.793 & 0.419\\
\ding{53} & \ding{51} & \ding{53} & \ding{53} & 0.876 & 0.497\\ 
\ding{53} & \ding{51} & \ding{53} & \ding{51} & 0.853 & 0.487\\
\ding{53} & \ding{51} & \ding{51} & \ding{53} & 0.853 & 0.478\\
\ding{53} & \ding{51} & \ding{51} & \ding{51} & 0.822 & 0.467\\

\bottomrule
\end{tabular}
\caption{Performance of the BRDialect ASR system on the processing settings of 5-gram KenLM Decoding (KD), Noise Cancellation (NC), Unicode Normalization (UN), and Punctuation Removal (PR).} 
\label{tab:tab_impact_brdialect}
\end{table}

\subsection{Processing Times of ASR Systems}
\label{sec:processing_time_and_transcription}


The inference time for audio files varies across different ASR systems, as shown in \cref{tab:inference_time}. The Whisper model exhibits significantly slower processing compared to BRDialect and IndicWav2Vec. Although the IndicWav2Vec model is slightly faster than BRDialect, its overall performance is considerably lower. In contrast, the processing time of BRDialect falls within an acceptable range for real-time communication applications.





 \label{sec:app_asr}

\section{TTS Systems Evaluation}
\label{sec:app_tts_eval}

In this study, we conduct extensive experiments with three open-source TTS systems, all of which are based on the VITS architecture \citep{kim2021conditional}: MMS-TTS-Ben \citep{pratap2024scaling}, and the VITS-Bengali Male and Female variants \citep{ComprehensiveBanglaTTS2023}. To evaluate the quality of the synthesized speech, we employ the mean opinion score (MOS) metric.

For a robust comparison, we curate a diverse set of texts from the BanSpeech dataset \citep{samin2024banspeech}, which contains audio-text pairs across thirteen categories. From each category, one representative text sample is selected, forming the test dataset, as detailed in \cref{sec:tts_sample_time}. Each text sample is synthesized using all three TTS models. To assess naturalness and perceived audio quality, an experienced human rater with expertise in audio signal processing independently rates each generated audio on a 1 to 5 scale, where 1 indicates very poor quality and 5 represents excellent quality \citep{streijl2016mean}.

\begin{figure*}[!t]
    \centering
    \includegraphics[width=\linewidth]{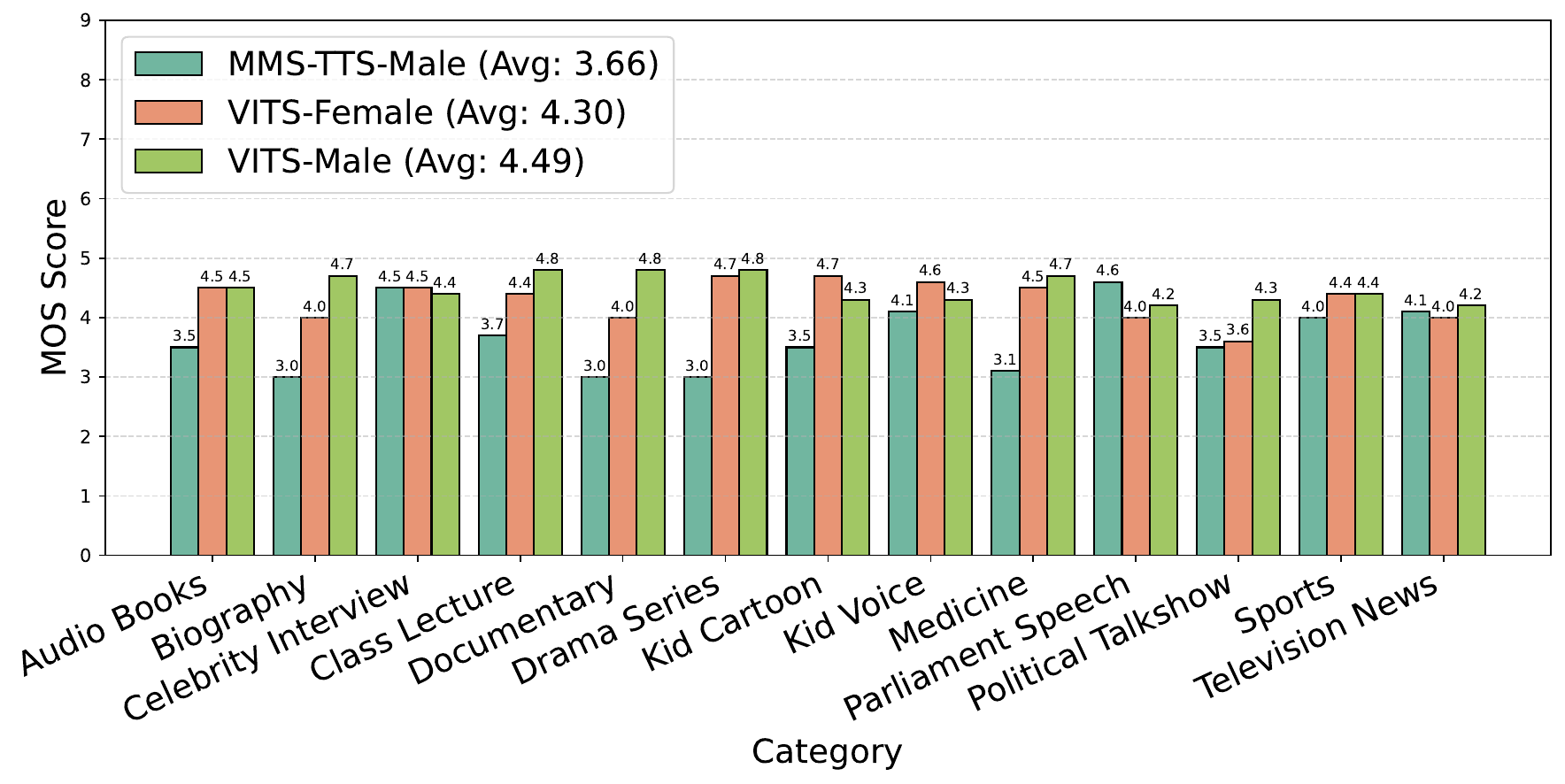}
    \caption{Mean Opinion Score (MOS) of three TTS models on the subset of the BanSpeech dataset.}
    \label{fig:mos_score}
\end{figure*}

The results, summarized in \cref{fig:mos_score}, indicate that the VITS-Bengali Male model achieves the highest average MOS score of 4.49, producing speech that is perceived as highly natural and pleasant. The VITS-Bengali Female model achieves an average MOS of 4.40, which is also suitable for end-to-end speech assistant systems. In contrast, MMS-TTS-Ben performs the lowest, with an average MOS score of 3.66, approximately 22.7\% lower than VITS-Bengali Male, indicating reduced suitability for end-to-end applications.

\begin{table}[!t]
\centering
\begin{tabular}{@{}lccc@{}}
\toprule
\multirow{2}{*}{\textbf{Audio}} & \multicolumn{3}{c}{\textbf{Preprocessing Times (s) $\downarrow$}} \\
\cmidrule{2-4}
& Whisper & IndicWav2Vec & BRDialect\\

\cmidrule{1-4}

sylhet\_1 & 3.07 & 0.65 & \textbf{0.77} \\
sylhet\_2 & 3.61 & 0.59 & \textbf{1.00} \\
sylhet\_3 & 3.99 & 1.36 & \textbf{1.31} \\
sylhet\_4 & 3.49 & 0.80 & \textbf{0.97} \\
\bottomrule

\end{tabular}
\caption{Processing time analysis of four audio files with an average duration of 8.75 s from the Sylhet region.} 
\label{tab:inference_time}
\end{table}

\subsection{Text Samples from the BanSpeech Dataset}
\label{sec:tts_sample_time}

\begin{figure*}[!h]
    \centering
    \includegraphics[width=\linewidth]{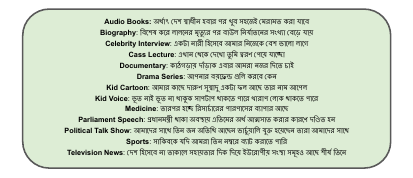}
    \caption{Selected text samples from the BanSpeech dataset to evaluate the performance of the TTS systems by calculating MOS scores.}
    \label{fig:text_sample_banspeech}
\end{figure*}

The BanSpeech dataset \citep{samin2024banspeech} comprises audio–text pairs spanning diverse categories. For evaluating the TTS systems, we use the text samples from this dataset, as illustrated in \cref{fig:text_sample_banspeech}.

 \label{sec:app_tts}

\section{End-to-End delay of BanglaTalk System}
\label{sec:end_to_end_delay}

To quantify the end-to-end delay of the BanglaTalk system, the delay for ten user queries is measured. \cref{tab:dealy_analysis} shows the results of this experiment. The BanglaTalk system has a low end-to-end delay of 4.9 s.

\begin{table}[!t]
\centering

\begin{tabular}{@{}lc@{}}
\toprule
\textbf{User Query} & \textbf{End-to-End Delay (s) $\downarrow$}\\
\cmidrule{1-2}
Query\_1 & 5 \\
Query\_2 & 4 \\
Query\_3 & 5 \\
Query\_4 & 5 \\
Query\_5 & 4 \\

Query\_6 & 6 \\
Query\_7 & 5 \\
Query\_8 & 5 \\
Query\_9 & 5 \\
Query\_10 & 5 \\
\cmidrule{1-2}
\textbf{Average} & \textbf{4.9}\\
\bottomrule

\end{tabular}
\caption{End-to-end delay analysis of the BanglaTalk system. Delay is calculated from the end of the user query to the start time of getting the audio response from the speech assistant.}
\label{tab:dealy_analysis}
\end{table} \label{sec:app_delay}

\section{User Study}
\label{sec:system_evaluation}

\begin{figure*}[!h]
    \centering
    \includegraphics[width=\linewidth]{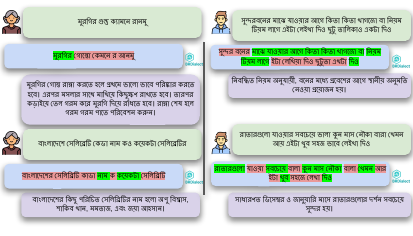}

    \caption{Conversational interactions with the BanglaTalk speech assistant using the regional dialects of Mymensingh (left) and Sylhet (right). Each example includes the user’s query, the transcript generated by the BRDialect ASR system, and the corresponding LLM-generated response. \textbf{Although the WER of BRDialect is relatively high in some cases, the LLM effectively captures the context and produces appropriate responses.}}
    
    \label{fig:interaction}
\end{figure*}

To evaluate the user experience with the BanglaTalk system, a rigorous qualitative analysis is performed. \cref{tab:rating} presents a summary of the experimental results, while \cref{fig:interaction} illustrates user interactions with the BanglaTalk system from two regions of Bangladesh -- Sylhet and Mymensingh.





\begin{table*}[!h]
\centering
\begin{tabular}{@{}llcccccc@{}}
\toprule
\multirow{2}{*}{\textbf{User}} & \multirow{2}{*}{\textbf{Region}} & \multicolumn{5}{c}{\textbf{Rating}} & \multirow{2}{*}{\textbf{Mean $\uparrow$}} \\
\cmidrule{3-7}
& & Query\_1 & Query\_2 & Query\_3 & Query\_4 & Query\_5 & \\

\cmidrule{1-8}
User\_1 & Sylhet & 3.0 & 4.0 & 2.5 & 3.0 & 5 & 3.5\\
User\_3 & Sylhet & 3.4 & 3.7 & 3.2 & 3.3 & 4.0 & 3.52 \\ 
User\_2 & Mymensingh & 2.0 & 4.0 & 3.8 & 4.2 & 3.3 & 3.46\\
User\_4 & Mymensingh & 5.0 & 3.0 & 4.5 & 3.5 & 4.0 & 4.0 \\
\cmidrule{1-8}

\textbf{Mean} & & & & & & & \textbf{3.62}\\
\bottomrule

\end{tabular}
\caption{User ratings on a {1–5} scale (where {1} indicates a poor experience and {5} indicates an excellent experience) for conversational interactions with the BanglaTalk speech assistant. Ratings are collected from four users representing two regional dialects -- Sylhet and Mymensingh.}
\label{tab:rating}
\end{table*}


\label{sec:appendix}

\end{document}